\documentclass[letterpaper, 10 pt, journal, twoside]{IEEEtran}
\usepackage{epsfig} 
\usepackage{amsmath} 
\usepackage{amssymb}  
\usepackage{color}
\usepackage{cite}
\usepackage[linesnumbered,ruled]{algorithm2e}
\usepackage[normalem]{ulem} 
\makeatletter
\let\NAT@parse\undefined
\makeatother
\usepackage{hyperref}
\hypersetup{pdfstartview=FitH,
            colorlinks=true,
            linkcolor=red,
            anchorcolor=blue,
            citecolor=black
            }
\usepackage{booktabs}
\usepackage{arydshln}
\usepackage{multirow}
\usepackage{mathrsfs}
\usepackage[caption=false,font=footnotesize]{subfig}
\usepackage[dvipsnames]{xcolor}

\usepackage{soul}
\usepackage{ulem}
\newcounter{RNum}
\renewcommand{\theRNum}{\arabic{RNum}}
\newcommand{\Remark}{\noindent\textit{\textbf{Remark}~\refstepcounter{RNum}\textbf{\theRNum}: }}
\newcommand{\NoOne}[1]{\textcolor{red}{#1}}
\newcommand{\NoTwo}[1]{\textcolor{green}{#1}}
\newcommand{\NoThree}[1]{\textcolor{blue}{#1}}
\usepackage{xcolor}
\soulregister\NoOne7
\soulregister\NoTwo7
\soulregister\NoThree7
\soulregister\Remark7

\soulregister{\cite}7 
\soulregister{\citep}7 
\soulregister{\citet}7 
\soulregister{\ref}7
\soulregister{\pageref}7 

\usepackage[absolute,overlay]{textpos}
\usepackage{multirow}
\usepackage{multicol}
\usepackage{graphicx}
\usepackage{amsmath}
\usepackage{amssymb}
\usepackage{booktabs}
\usepackage{xcolor,colortbl}
\usepackage{makecell}
\usepackage{pifont}
\usepackage{cancel}
\usepackage{adjustbox}

\newcommand{\eg}{\textit{e.g.}}
\newcommand{\ie}{\textit{i.e.}}

\definecolor{nbarrier}{RGB}{255, 120, 50}
\definecolor{nbicycle}{RGB}{255, 192, 203}
\definecolor{nbus}{RGB}{255, 255, 0}
\definecolor{ncar}{RGB}{0, 150, 245}
\definecolor{nconstruct}{RGB}{0, 255, 255}
\definecolor{nmotor}{RGB}{200, 180, 0}
\definecolor{npedestrian}{RGB}{255, 0, 0}
\definecolor{ntraffic}{RGB}{255, 240, 150}
\definecolor{ntrailer}{RGB}{135, 60, 0}
\definecolor{ntruck}{RGB}{160, 32, 240}
\definecolor{ndriveable}{RGB}{255, 0, 255}
\definecolor{nother}{RGB}{139, 137, 137}
\definecolor{nsidewalk}{RGB}{75, 0, 75}
\definecolor{nterrain}{RGB}{150, 240, 80}
\definecolor{nmanmade}{RGB}{213, 213, 213}
\definecolor{nvegetation}{RGB}{0, 175, 0}

\definecolor{car}{rgb}{0.39215686, 0.58823529, 0.96078431}
\definecolor{bicycle}{rgb}{0.39215686, 0.90196078, 0.96078431}
\definecolor{motorcycle}{rgb}{0.11764706, 0.23529412, 0.58823529}
\definecolor{truck}{rgb}{0.31372549, 0.11764706, 0.70588235}
\definecolor{other-vehicle}{rgb}{0.39215686, 0.31372549, 0.98039216}
\definecolor{person}{rgb}{1.        , 0.11764706, 0.11764706}
\definecolor{bicyclist}{rgb}{1.        , 0.15686275, 0.78431373}
\definecolor{motorcyclist}{rgb}{0.58823529, 0.11764706, 0.35294118}
\definecolor{road}{rgb}{1.        , 0.        , 1.        }
\definecolor{parking}{rgb}{1.        , 0.58823529, 1.        }
\definecolor{sidewalk}{rgb}{0.29411765, 0.        , 0.29411765}
\definecolor{other-ground}{rgb}{0.68627451, 0.        , 0.29411765}
\definecolor{building}{rgb}{1.        , 0.78431373, 0.        }
\definecolor{fence}{rgb}{1.        , 0.47058824, 0.19607843}
\definecolor{vegetation}{rgb}{0.        , 0.68627451, 0.        }
\definecolor{trunk}{rgb}{0.52941176, 0.23529412, 0.        }
\definecolor{terrain}{rgb}{0.58823529, 0.94117647, 0.31372549}
\definecolor{pole}{rgb}{1.        , 0.94117647, 0.58823529}
\definecolor{traffic-sign}{rgb}{1.        , 0.        , 0.    }    

\makeatletter
\newcommand{\car@semkitfreq}{3.92}
\newcommand{\bicycle@semkitfreq}{0.03}
\newcommand{\motorcycle@semkitfreq}{0.03}
\newcommand{\truck@semkitfreq}{0.16}
\newcommand{\othervehicle@semkitfreq}{0.20}
\newcommand{\person@semkitfreq}{0.07}
\newcommand{\bicyclist@semkitfreq}{0.07}
\newcommand{\motorcyclist@semkitfreq}{0.05}
\newcommand{\road@semkitfreq}{15.30}  %
\newcommand{\parking@semkitfreq}{1.12}
\newcommand{\sidewalk@semkitfreq}{11.13}  %
\newcommand{\otherground@semkitfreq}{0.56}
\newcommand{\building@semkitfreq}{14.1}  %
\newcommand{\fence@semkitfreq}{3.90}
\newcommand{\vegetation@semkitfreq}{39.3}  %
\newcommand{\trunk@semkitfreq}{0.51}
\newcommand{\terrain@semkitfreq}{9.17} %
\newcommand{\pole@semkitfreq}{0.29}
\newcommand{\trafficsign@semkitfreq}{0.08}
\newcommand{\semkitfreq}[1]{{\csname #1@semkitfreq\endcsname}}

\ifCLASSINFOpdf
\else
\fi
\begin{document}
%
\title{Co-Occ: Coupling Explicit Feature Fusion with Volume Rendering Regularization for Multi-Modal 3D Semantic Occupancy Prediction}
%
%
%

\author{Jingyi Pan$^{1}$, Zipeng Wang$^{1}$, Lin Wang$^{1, 2, *}$ 
\thanks{Manuscript received: February, 8, 2024; Revised March, 18, 2024; Accepted April, 21, 2024.}
\thanks{This paper was recommended for publication by Associate Editor R. Boutteau and Editor P. Vasseur upon evaluation of the Associate Editor and Reviewers' comments. This work was supported by the Guangzhou Fundamental and Applied Basic Research under Grant No.2024A04J4072.} 
\thanks{$^{*}$Corresponding author}
\thanks{$^{1}$Jingyi Pan and Zipeng Wang are with the AI Thrust, The Hong Kong University of Science and Technology (Guangzhou), Guangdong 511458, China.
        {\tt\small \{jpan305, zwang253\}@connect.hkust-gz.edu.cn}}%
\thanks{$^{1,2}$Lin Wang is with AI/CMA Thrust, HKUST(GZ) and Dept. of CSE, HKUST, Hong Kong SAR, China, Email: {\tt\small linwang@ust.hk}}
\thanks{Digital Object Identifier (DOI): 10.1109/LRA.2024.3396092}
}

\maketitle

\begin{abstract}
3D semantic occupancy prediction is a pivotal task in the field of autonomous driving. 
Recent approaches have made great advances in 3D semantic occupancy predictions on a single modality.
However, multi-modal semantic occupancy prediction approaches have encountered difficulties in dealing with the modality heterogeneity, modality misalignment, and insufficient modality interactions that arise during the fusion of different modalities data, which may result in the loss of important geometric and semantic information.
This letter presents a novel multi-modal, \ie, LiDAR-camera 3D semantic occupancy prediction framework, dubbed \textbf{Co-Occ}, which couples explicit LiDAR-camera feature fusion with implicit volume rendering regularization. 
The key insight is that volume rendering in the feature space can proficiently bridge the gap between 3D LiDAR sweeps and 2D images while serving as a physical regularization to enhance LiDAR-camera fused volumetric representation.
Specifically, we first propose a Geometric- and Semantic-aware Fusion (GSFusion) module to explicitly enhance LiDAR features by incorporating neighboring camera features through a K-nearest neighbors (KNN) search.
Then, we employ volume rendering to project the fused feature back to the image planes for reconstructing color and depth maps. 
These maps are then supervised by input images from the camera and depth estimations derived from LiDAR, respectively.
Extensive experiments on the popular nuScenes and SemanticKITTI benchmarks verify the effectiveness of our Co-Occ for 3D semantic occupancy prediction.  
The project page is available at \url{https://rorisis.github.io/Co-Occ_project-page/}.
\end{abstract}


%
\IEEEpeerreviewmaketitle

\section{INTRODUCTION}

\IEEEPARstart{3D} semantic occupancy prediction is a task that involves estimating the geometric structure and semantic categories of occupied voxels in a scene simultaneously, which has been widely applied in robot manipulation~\cite{varley2017shape}, robot navigation~\cite{wang2021learning} and autonomous driving~\cite{yan2021sparse, mei2023ssc}. While earlier methods primarily focused on indoor environments and utilized dense geometric information from LiDAR or depth sensors, outdoor 3D semantic occupancy prediction relies on sparse point clouds and multi-view or monocular camera images. Besides, unlike other 3D perception tasks that predominantly focus on specific classes of foreground objects (\eg, 3D object detection), 3D semantic occupancy prediction requires a comprehensive perception of the surroundings by predicting the geometry and semantics within the scene~\cite{caesar2020nuscenes, behley2019semantickitti}. 

Leveraging the complementary strengths of LiDAR and camera data is crucial in various 3D perception tasks. Cameras provide rich semantic information but lack precise geometric details, while LiDAR offers accurate depth or spatial information but may suffer from sparse contextual details~\cite{liu2023bevfusion, li2022unifying, li2023mseg3d, qin2023supfusion}. Therefore, many 3D perception methods combine both modalities through camera-to-LiDAR or LiDAR-to-camera projection~\cite{huang2020epnet, li2023mseg3d}, unification in the voxel space~\cite{li2022unifying, wang2023openoccupancy} or in bird's eye view (BEV) space~\cite{liu2023bevfusion}.

However, the fusion of LiDAR-camera data for 3D semantic occupancy prediction is not a straightforward task due to the heterogeneity between the modalities and the limited interaction between them. 
Specifically, LiDAR sweeps capture sparse 3D points, whereas cameras capture dense color information on the image plane. 
Most existing methods~\cite{li2022unifying, wang2023openoccupancy} merge LiDAR and camera features by elevating 2D image features to the 3D voxel space in a pixel-to-point fashion.
However, due to the extrinsic calibration inaccuracies between LiDAR and camera~\cite{chen2020novel}, the geometrically-aware 2D to 3D view transformation~\cite{philion2020lift} might not efficiently elevate the 2D feature to the corresponding point in the 3D space. Moreover, the inaccurate fused volumetric representations lead to inconsistent occupancy predictions and loss of semantic information.

\begin{figure}[t!]
    \centering
    \includegraphics[width=\linewidth]{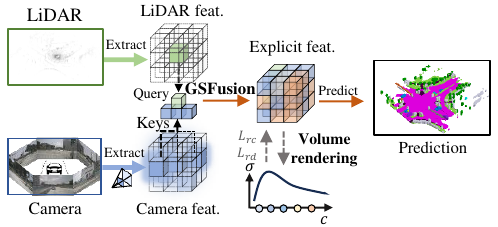}
    \vspace{-20pt}
    \caption{\textbf{The pipeline of our Co-Occ}. Our method utilizes the GSFusion module to acquire explicit fused features that retain both the semantic benefits from the cameras and the geometric benefits from LiDAR. Then, implicit volume rendering-based regularization is applied to bridge the gap between 3D LiDAR and 2D images and enhance fused representation.}
    \vspace{-15pt}
    \label{fig:1}
\end{figure}

In this letter, we present \textbf{Co-Occ}, a novel multi-modal 3D semantic occupancy prediction framework. 
Co-Occ couples explicit 3D feature representation with implicit volume rendering-based regularization to strengthen inter-modal interaction and enhance the fused volumetric representation, as shown in Fig.~\ref{fig:1}. 
Firstly, we extract features from LiDAR and camera data and project them into a unified voxel space. We then fuse the features via a Geometric- and Semantic-aware Fusion (GSFusion) module, which enhances LiDAR features with neighboring camera features within the geometric-aligned voxel space. 
To accomplish this, we use a K-nearest neighbors (KNN) search to identify relevant camera features, which are further selected using a KNN gate operation. 
The GSFusion module explicitly incorporates the semantic information from camera features into LiDAR features, particularly for the sparse input.

Then, we incorporate implicit volume rendering-based regularization to supervise the fused explicit representations. Inspired by recent advancements in neural rendering~\cite{mildenhall2021nerf}, we cast rays from the camera into the scene and sample along these rays uniformly. The corresponding features of these samples are retrieved from the fused feature, and two auxiliary heads predict the density and color of these samples.  
The color and depth are then projected back onto the 2D image plane and supervised by ground truth of color from cameras and depth maps from LiDAR. 
\textit{This enables us to effectively bridge the gap between 3D LiDAR sweeps and 2D camera images and enhance the fused volumetric representation}. 
We then employ the fused LiDAR-camera features for decoding and occupancy prediction. It's worth noting that the volume rendering regularization is only applied during training and does not impact the inference time.

We conducted extensive experiments on the nuScenes~\cite{caesar2020nuscenes} and SemanticKITTI~\cite{behley2019semantickitti} benchmarks. 
The results demonstrate that our Co-Occ effectively boosts the accuracy and density of semantic occupancy prediction. 
We establish the new state-of-the-art, with \textbf{41.1\%} IoU and \textbf{27.1\%} mIoU on the nuScenes validation set with a voxel size of [0.5m, 0.5m, 0.5m]~\cite{caesar2020nuscenes, wei2023surroundocc}. Also, on the SemanticKITTI test set, we achieve \textbf{56.6\%} IoU and \textbf{24.4\%} mIoU.
To sum up, our major contributions are three-fold: (\textbf{I}) We propose a novel multi-modal 3D semantic occupancy prediction framework as it can efficiently take LiDAR and camera inputs; (\textbf{II}) We propose the GSFusion module together with the implicit volume rendering-based regularization. This optimally leverages each modality while ensuring consistent, fine-grained unified volumetric representation; (\textbf{III}) Extensive experiments on the nuScenes datasets and the SemanticKITTI benchmark demonstrate the superiority of our method.



\begin{figure*}[t]
    \centering
    \includegraphics[width=\linewidth]{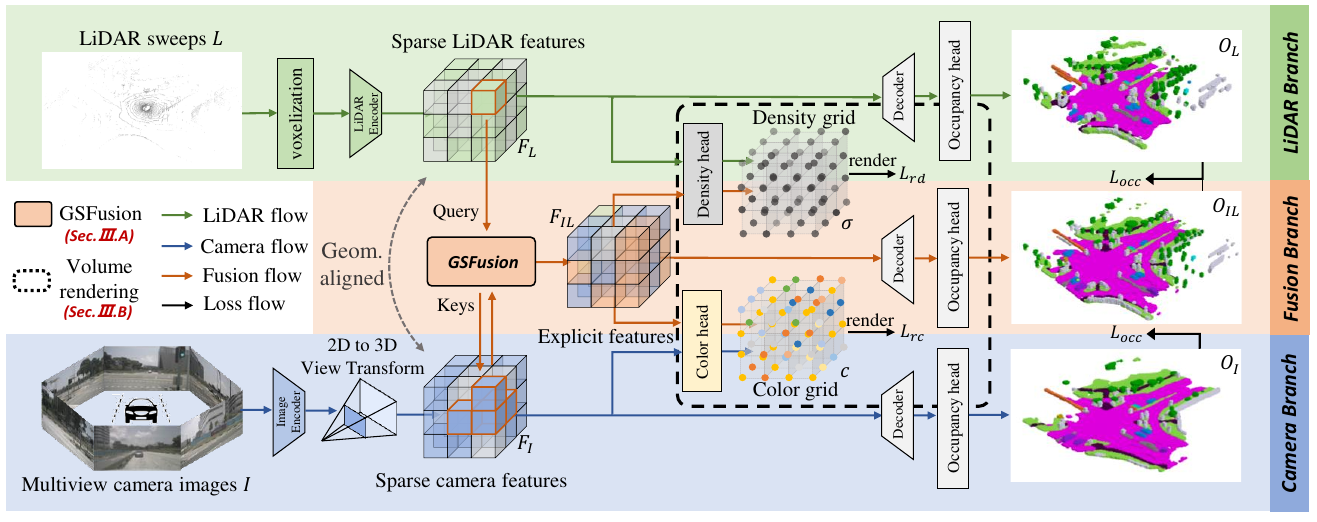}
    \vspace{-18pt}
    \caption{\textbf{Our Co-Occ framework}. It consists of an explicit GSFusion module and implicit volume rendering regularization. The GSFusion module (Fig.~\ref{fig:gsfusion}) takes advantage of both the semantic benefits derived from camera features and the geometric benefits obtained from LiDAR. Meanwhile, the implicit volume rendering regularization (Fig.~\ref{fig:imp_render}) guarantees the fusion of explicit LiDAR-camera features in an accurate and detailed manner, which further enhances the performance of 3D semantic prediction. Notably, implicit volume rendering regularization is only utilized during the training process.}
    \vspace{-10pt}
    \label{fw}
\end{figure*}

\section{RELATED WORK}

\noindent\textbf{3D Semantic Occupancy Prediction.}
The objective of 3D semantic occupancy prediction is to estimate the surrounding scene by incorporating both geometry and semantics.
Earlier works primarily focused on indoor scenarios~\cite{song2017semantic, zhang2018efficient, liu2018see}.
In recent years, S3CNet~\cite{cheng2021s3cnet}, JS3C-Net~\cite{yan2021sparse}, and SSA-SC~\cite{yang2021semantic} refine predictions using sparse point clouds with different representations in outdoor scenarios. MonoScene~\cite{cao2022monoscene} is the first work using only RGB inputs. Recent advancements include SCPNet~\cite{xia2023scpnet}, OccFormer~\cite{zhang2023occformer}, and SurroundOcc~\cite{wei2023surroundocc}, which use multi-path networks to aggregate multi-scale features for semantic occupancy prediction. Approaches like \cite{pan2023uniocc, pan2023renderocc} directly predict 3D semantic occupancy using NeRF~\cite{mildenhall2021nerf}, but rendering speed limits their efficiency. Openoccupancy~\cite{wang2023openoccupancy} introduces a benchmark for LiDAR-Camera semantic occupancy prediction. In this letter, we enhance the performance of 3D semantic occupancy prediction through explicit feature fusion and implicit volume rendering regularization.



\noindent\textbf{LiDAR-Camera Fusion-Based 3D Perception.}
LiDAR and camera sensors are widely used in various 3D perception tasks. Existing research has developed different LiDAR-camera fusion techniques for tasks like 3D object detection~\cite{liu2023bevfusion, li2022deepfusion, zhou2023sparsefusion} and LiDAR segmentation~\cite{zhang2023lidar, li2023mseg3d}. Fusion methods can be categorized as: \textbf{1)} Project-based fusion~\cite{huang2020epnet, li2023mseg3d}, combining image features with raw LiDAR points or projecting LiDAR point clouds into a range view and fusing with image features. \textbf{2)} Feature-level fusion~\cite{liu2023bevfusion, zhang2023lidar, wang2019frustum}, projecting LiDAR and image data into a shared feature space like BEV or voxels. \textbf{3)} Attention-based fusion~\cite{bai2022transfusion, li2022deepfusion}, utilizing LiDAR features as proposals to query image features through cross-attention. 
In contrast to implicit fusion methods designed for specific objects or sparse 3D perception, our method initially leverages KNN-based fusion methods to expand the semantic perception field with aligned geometric representations of two modalities, reducing errors from inaccurate calibrations. Additionally, we utilize volume rendering to bridge the gap between 2D and 3D representations and enhance the volumetric fused features across different modalities.

\noindent\textbf{Scene Understanding with Volume Rendering.}
Recently, 3D perception approaches inspired by Neural Radiance Field (NeRF)~\cite{mildenhall2021nerf} have employed volume rendering to generate 3D scene representations or estimate 3D geometry information from multi-view images. 
Previous studies, such as \cite{kundu2022panoptic, wu2023mars}, have focused on generating implicit scene representations through per-scene optimization using multi-view images. However, this approach can pose challenges when it comes to generalizing the results to different scenes.
Approaches like \cite{pan2023uniocc, pan2023renderocc, zhang2023occnerf} utilize NeRF to directly obtain 3D semantic occupancy predictions or acquire the rendered 2D semantic to supervise the semantic information. 
Unlike previous methods, we propose an implicit volume rendering-based regularization to enforce regularization on the fusion of LiDAR and camera data, based on the RGB and depth, to further enhance the unified volumetric representation.

\section{PROPOSED METHOD}
Our objective is to predict the 3D occupancy of surrounding scenes using LiDAR sweeps, denoted as $L$, and its corresponding surround-view images, represented as $I=\{I_1, ..., I_N\}$, where $N$ indicates the total number of camera views in a scene. Our Co-Occ framework comprises two key components: an explicit Geometric- and Semantic-Aware fusion module (GSFusion, Sec.~\ref{fusion}), and an implicit volume rendering -based regularization achieved in the feature space (Sec.~\ref{rendering}). We provide a detailed description of the optimization of our framework in Sec.~\ref{loss}.

\subsection{Geometric- and Semantic-aware Fusion} 
\label{fusion}

\begin{figure}[t!]
    \centering
    \includegraphics[width=\linewidth]{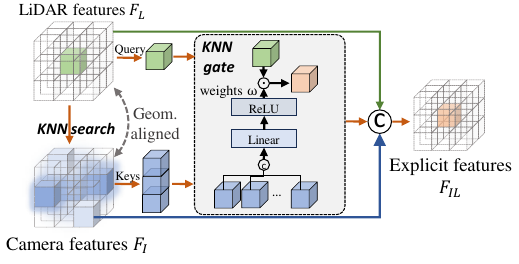}
    \vspace{-18pt}
    \caption{The workflow of the GSFusion module begins with searching for $K$ nearest neighbors from camera features to supplement the semantic context of LiDAR features. A KNN gate is then used to obtain weights to boost the LiDAR features. The final step involves concatenating the features.}
    \vspace{-15pt}
    \label{fig:gsfusion}
\end{figure}


Directly projecting images into LiDAR space~\cite{li2022unifying, bai2022transfusion, song2023graphalign} may introduce errors due to inaccurate extrinsic calibration and depth prediction~\cite{chen2020novel}. Also, a substantial amount of empty voxel space may impede the interaction between the two modalities, leading to inconsistent occupancy predictions and loss of semantic information. To overcome these challenges, we propose GSFusion, which incorporates a KNN search and KNN gate, as illustrated in Fig.~\ref{fig:gsfusion}.

\noindent\textbf{KNN Search.}
We employ a 2D image encoder~\cite{he2016deep, liu2021swin, lin2017feature} to derive camera features, followed by a 2D-to-3D view transformation~\cite{philion2020lift, wang2023openoccupancy} to project these 2D features into 3D sparse image features $F_I$ as voxel representation. 
Simultaneously, we encode the LiDAR sweep into a 3D sparse feature $F_L$ utilizing voxelization and 3D LiDAR encoder~\cite{yan2018second}. Consequently, the camera features $F_I$ and LiDAR features $F_L$ are aligned with the same dimension ${\mathbb{R}^{D \times H \times W \times{C}}}$, where $C$ signifies the feature dimensions.

We then select the 3D coordinates of non-empty LiDAR and camera features, denoted as $P_L \in \mathbb{R}^{N_L\times3}$ and $P_I \in \mathbb{R}^{N_I\times3}$. Here, $N_L$ and $N_I$ represent the quantity of non-empty features, and each entry corresponds to a 3D coordinate $(x,y,z)$. 
Due to $F_I$ and $F_L$ being geometrically aligned, we can directly search $K$ nearest neighbors of a given LiDAR coordinate in the voxel space within a specific radius $r$.
For the $i$-th non-empty LiDAR feature, we denote its neighboring color features as $\{N_i^0, N_i^1, ..., N_i^k\}$, where $k$ is the hyper-parameter of the neighboring number.
We implement a fast KNN search algorithm with CUDA kernel, which ensures the efficiency of KNN search.

\noindent\textbf{KNN Gate.}
After obtaining the $K$ nearest neighborhood coordinates from the camera features, we propose a learnable KNN gate (see Fig.~\ref{fig:gsfusion}) to obtain the semantic weight $\omega_i$ of each $i$-th non-empty LiDAR feature for further fusion with LiDAR features.
{\setlength\abovedisplayskip{2pt}
\setlength\belowdisplayskip{2pt}
\begin{equation}
    \omega_i = {\rm Linear}({\rm Concat}(N_i^0, N_i^1, ..., N_i^k)),
\end{equation}
And we derive the fused LiDAR-camera features $F_{IL}$ as:
{\setlength\abovedisplayskip{2pt}
\setlength\belowdisplayskip{2pt}
\begin{equation}
    F_{IL} = {\rm Concat}(F_I, F_L, F_L\cdot \omega),
\end{equation}}
where $\omega$ represents the total semantic weight of the LiDAR features from $w_i$.
\vspace{-5pt}
\subsection{Implicit Volume Rendering Regularization}
\label{rendering}
After fusing LiDAR-camera features using the GSFusion module, we propose a volume rendering-based regularization method depicted in Fig.~\ref{fig:imp_render}. Unlike existing scene understanding methods that focus on processing data solely from images or points using volume rendering techniques\cite{siddiqui2023panoptic,xu2023mononerd, pan2023renderocc}, our approach applies volume rendering in a generalized manner to regulate explicit features fused from LiDAR and camera data across different scenes, instead of a per-scene basis. This allows us to impose a physical constraint and promote consistency within the fused features.

\begin{figure}[t!]
    \centering
    \includegraphics[width=\linewidth]{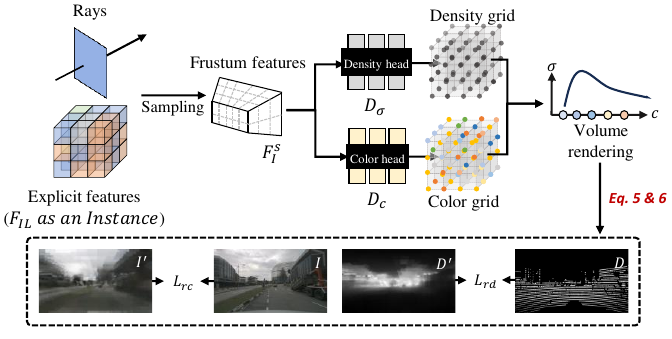}
    \vspace{-23pt}
    \caption{The implicit volume rendering-based regularization involves obtaining frustum features from rays and explicit features. Frustum features are used to create the density grid and color grid, which are then utilized to generate the depth map and color map.}
    \label{fig:imp_render}
    \vspace{-15pt}
\end{figure}

\noindent\textbf{Sampling along Rays in Feature Space.}
Following neural rendering techniques~\cite{mildenhall2021nerf}, we cast rays from the camera into the space and uniformly sample those rays, leading to 
3D coordinates $X \in \mathbb{R}^{N\times n_s\times h \times w \times 3}$ in ego-car space. Here, $n_s$ denotes the number of sampling points on each ray and $h, w$ denotes the height and width of the rendered image size. To reduce the computational cost, we set $h$ and $w$ to be $1/16$ of the original image.
We then use trilinear interpolation to generate a frustum feature $F_I^s$ corresponding the sampled coordinates $X$ from the fused feature $F_{IL}$:
{\setlength\abovedisplayskip{2pt}
\setlength\belowdisplayskip{2pt}
\begin{equation}
    \label{eq:sample}
    F_I^s = {\rm Interpolate}(F_{IL}, X)
\end{equation}
where $F_I^s \in \mathbb{R}^{N\times n_s\times h \times w \times C}$, and ${\rm Interpolate(*)}$ queries the feature $F_{IL}$ at location $X$.

We employ two Multi-Layer Perceptron (MLP) networks, $D_{c}$ and $D_{\sigma}$, to generate color grid $c$ and density grid $\sigma$ of points respectively:
{\setlength\abovedisplayskip{2pt}
\setlength\belowdisplayskip{2pt}
\begin{equation}
\begin{split}
    c &= {\rm Sigmoid}(D_c(F_I^s)),\\
    \sigma &= {\rm ReLU}(D_\sigma(F_I^s))
\end{split}
\end{equation}
where $c \in \mathbb{R}^{N\times n_s\times h \times w \times 3}$ and $\sigma \in \mathbb{R}^{N\times n_s\times h \times w \times 1}$.

\noindent\textbf{Volume Rendering for Depth and Color.}
We render the predicted color map $\hat{I}$ using volume rendering:
{\setlength\abovedisplayskip{2pt}
\setlength\belowdisplayskip{2pt}
\begin{equation}
\begin{split}
    &\hat{I} = \sum_{i=1}^{n_s}T_i(1-{\rm exp(-\sigma_i\delta_{t_{i}})})c_i, 
\end{split}
\end{equation}
where $T_i = {\rm exp}(-\sum_{j=1}^{i-1}\sigma_j\delta_{t_{j}})$ and $\delta_t$ is the distance between two sampled points.
Similarly, we obtain the depth map $\hat{D}$ from the density field:
{\setlength\abovedisplayskip{2pt}
\setlength\belowdisplayskip{2pt}
\begin{equation}
    \hat{D} = \sum_{i=1}^{n_s}T_i(1 - {\rm exp}(-\sigma_i\delta_{t_{i}})).
\end{equation}
 We then upsample $\hat{I}$ and $\hat{D}$ to image resolution as $\{\hat{I'}, \hat{D'}\} \in \{\mathbb{R}^{h'\times w'\times3}, \mathbb{R}^{h'\times w'\times1}\}$. Here, $h'$ and $w'$ represent the height and the width of the image resolution.
Such rendered RGB images and depth are supervised by color loss, depth loss, and opacity loss with input camera and LiDAR in Sec.~\ref{loss}.

\begin{table*}[t!]
\setlength{\tabcolsep}{0.0035\linewidth}
\newcommand{\classfreq}[1]{{~\tiny(\semkitfreq{#1}\%)}}  %
\centering
\caption{\textbf{3D Semantic occupancy prediction results on nuScenes validation set.} We report the geometric metric IoU, semantic metric mIoU, and the IoU for each semantic class. The C and L denotes camera and LiDAR, respectively. }
\vspace{-3mm}
\resizebox{.85\linewidth}{!}{
\begin{tabular}{l|c | c c | c c c c c c c c c c c c c c c c|c c}
    \toprule
    Method
    & \makecell[c]{Modality} 
    & IoU& mIoU
    & \rotatebox{90}{\textcolor{nbarrier}{$\blacksquare$} barrier}
    & \rotatebox{90}{\textcolor{nbicycle}{$\blacksquare$} bicycle}
    & \rotatebox{90}{\textcolor{nbus}{$\blacksquare$} bus}
    & \rotatebox{90}{\textcolor{ncar}{$\blacksquare$} car}
    & \rotatebox{90}{\textcolor{nconstruct}{$\blacksquare$} const. veh.}
    & \rotatebox{90}{\textcolor{nmotor}{$\blacksquare$} motorcycle}
    & \rotatebox{90}{\textcolor{npedestrian}{$\blacksquare$} pedestrian}
    & \rotatebox{90}{\textcolor{ntraffic}{$\blacksquare$} traffic cone}
    & \rotatebox{90}{\textcolor{ntrailer}{$\blacksquare$} trailer}
    & \rotatebox{90}{\textcolor{ntruck}{$\blacksquare$} truck}
    & \rotatebox{90}{\textcolor{ndriveable}{$\blacksquare$} drive. suf.}
    & \rotatebox{90}{\textcolor{nother}{$\blacksquare$} other flat}
    & \rotatebox{90}{\textcolor{nsidewalk}{$\blacksquare$} sidewalk}
    & \rotatebox{90}{\textcolor{nterrain}{$\blacksquare$} terrain}
    & \rotatebox{90}{\textcolor{nmanmade}{$\blacksquare$} manmade}
    & \rotatebox{90}{\textcolor{nvegetation}{$\blacksquare$} vegetation}& Input Size & 2D Backbone \\
    \midrule
    MonoScene~\cite{cao2022monoscene} & C  & 24.0 & 7.3 & 4.0  & 0.4  &  8.0 &  8.0 & 2.9  & 0.3  &  1.2& 0.7& 4.0 & 4.4 & 27.7 & 5.2  & 15.1& 11.3 & 9.0 & 14.9& $900\times 1600$& R101-DCN \\

        BEVFormer~\cite{li2022bevformer} &C&30.5& 16.7 &14.2& 6.5 &23.4& 28.2& 8.6& 10.7 &6.4& 4.0 &11.2& 17.7 &37.2& 18.0& 22.8& 22.1 &13.8& 22.2& $900\times 1600$& R101-DCN\\  
        SurroundOcc~\cite{wei2023surroundocc} & C & 31.4& 20.3 &20.5 &11.6 &28.1 &30.8& 10.7 &15.1 &14.0 &12.0 &14.3 &22.2& 37.2& 23.7& 24.4& 22.7& 14.8 &21.8& $900\times 1600$& R101-DCN\\ 
        OccFormer~\cite{zhang2023occformer} &C&29.9& 20.1 &21.1& 11.3 &28.2& 30.3& 10.6& 15.7 &14.4& 11.2 &14.0& 22.6 &37.3&22.4 & 24.9& 23.5 &15.2& 21.1& $896\times 1600$& R101\\
        C-CONet~\cite{wang2023openoccupancy} & C & 26.1& 18.4 &18.6 &10.0 &26.4 &27.4& 8.6 &15.7 &13.3 &9.7 &10.9 &20.2& 33.0& 20.7& 21.4& 21.8& 14.7 &21.3& $896\times 1600$& R101\\
       FB-Occ~\cite{li2023fb}& C & 31.5& 19.6& 20.6 &11.3 &26.9 &29.8&10.4& 13.6&13.7 &11.4 &11.5 &20.6&38.2& 21.5& 24.6& 22.7& 14.8& 21.6& $896\times 1600$& R101 \\
        RenderOcc~\cite{pan2023renderocc}& C & 29.2& 19.0 &19.7 &11.2 &28.1 &28.2& 9.8 &14.7 &11.8 &11.9&13.1 &20.1& 33.2& 21.3& 22.6& 22.3& 15.3 &20.9& $896\times 1600$& R101\\
        LMSCNet~\cite{roldao2020lmscnet} & L & 36.6 & 14.9 & 13.1&  4.5 & 14.7  & 22.1  & 12.6  &  4.2 & 7.2 & 7.1&  12.2&  11.5& 26.3 & 14.3  & 21.1 & 15.2  & 18.5 & 34.2& -& - \\

        L-CONet~\cite{wang2023openoccupancy} & L&  39.4&17.7 & 19.2 & 4.0& 15.1 & 26.9 & 6.2 & 3.8 &  6.8& 6.0& 14.1&  13.1& 39.7 & 19.1 & 24.0  & 23.9 & 25.1 & 35.7& -& - \\
        M-CONet~\cite{wang2023openoccupancy} & C\&L &  39.2&24.7  & 24.8 & 13.0& 31.6 & 34.8 & 14.6 & 18.0  &  20.0& 14.7& 20.0&  26.6& 39.2 & 22.8 & 26.1  & 26.0 & 26.0 & 37.1& $896\times 1600$ & R101 \\ \hline
        Co-Occ (Ours) & C &30.0& 20.3& 22.5& 11.2& 28.6& 29.5&  9.9& 15.8& 13.5 & 8.7&13.6& 22.2& 34.9& 23.1& 24.2&24.1  &18.0& 24.8& $896\times 1600$&R101 \\
        Co-Occ (Ours) & L&\textbf{42.2}&22.9 &22.0&6.90 & 25.7&32.4&  14.5& 13.5&\textbf{21.0 }&10.5 &18.0&22.5& 36.6& 21.8& 24.6&25.7 &\textbf{31.2}  & \textbf{39.9} &-&-\\
        Co-Occ (Ours) & C\&L&41.1&\textbf{27.1}  &\textbf{28.1} &\textbf{16.1} &\textbf{34.0} &\textbf{37.2} &\textbf{17.0}&\textbf{21.6} &20.8 &\textbf{15.9} &\textbf{21.9} &\textbf{28.7}&\textbf{42.3}&\textbf{25.4} &\textbf{29.1} &\textbf{28.6} &28.2 &38.0& $896\times 1600$&R101   \\
    \bottomrule

\end{tabular}}
\label{tab:nusc}
\end{table*}

\begin{figure*}[t!]
    \centering
    \vspace{-10pt}
    \includegraphics[width=0.9\linewidth]{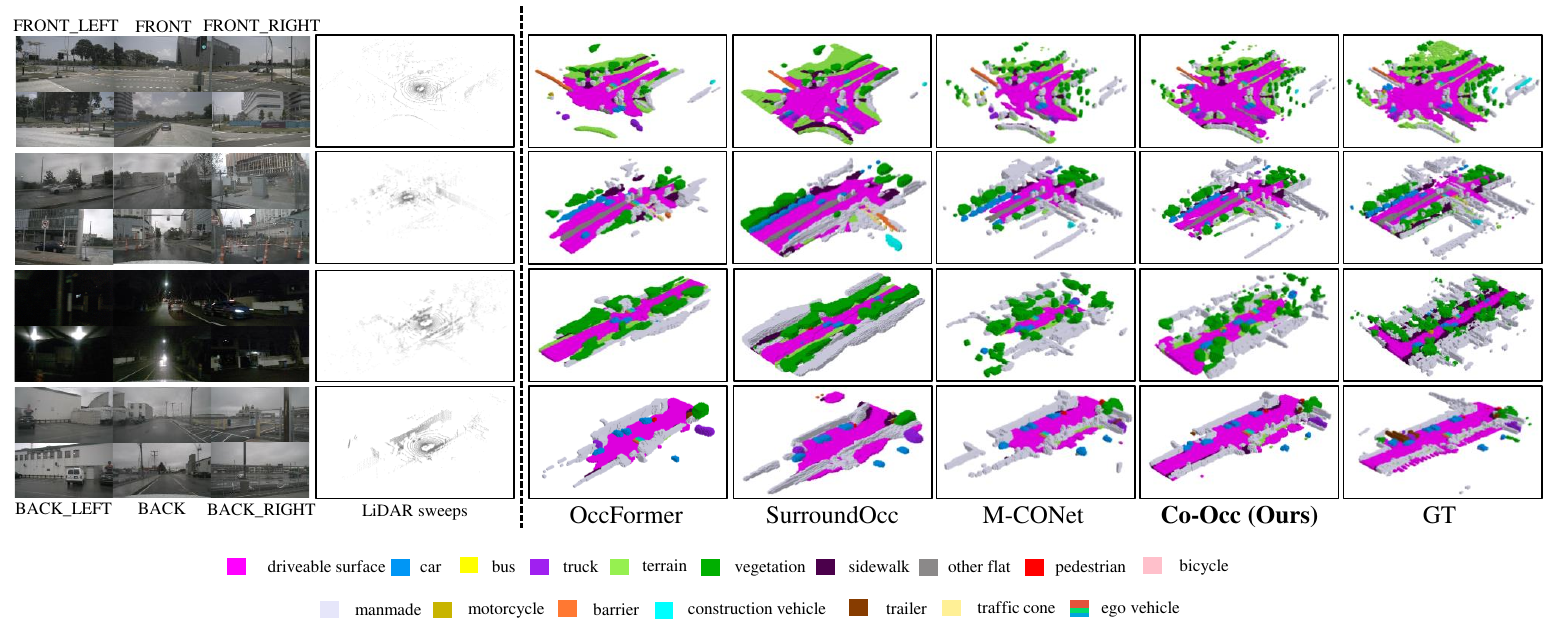}
    \vspace{-10pt}
    \caption{The qualitative comparisons results on nuScenes validation set. The leftmost column shows the input surrounding images and LiDAR sweeps, the following three columns visualize the 3D semantic occupancy prediction from OccFormer~\cite{zhang2023occformer}, SurroundOcc~\cite{wei2023surroundocc} (these two predicts results using only cameras), M-CONet~\cite{wang2023openoccupancy}, our Co-Occ, and the annotation from~\cite{wei2023surroundocc}. \textbf{Better viewed when zoomed in.}}
    \vspace{-10pt}
    \label{fig:com_nusc}
\end{figure*}

\begin{table}[t!]
    \centering
    \footnotesize
    \caption{Ablation study on the impact of different components.}
    \vspace{-5pt}
    \setlength{\tabcolsep}{0.04\linewidth}
    \begin{tabular}{cccc|cc}
    \toprule
    \multirow{2}{*}{Base} & Fusion & \multicolumn{2}{c|}{Rendering}& \multirow{2}{*}{IoU} & \multirow{2}{*}{mIoU} \\ \cline{2-4}
     & GSFusion & $\mathcal{L}_{rc}$ & $\mathcal{L}_{rd}$ &  &  \\ \midrule
    \checkmark  & & & &38.2 &24.2\\
    \checkmark  &\checkmark & & &40.1 &25.9\\
    \checkmark  &\checkmark &\checkmark & &40.8 &26.4\\
    \checkmark  &\checkmark &\checkmark &\checkmark &\textbf{41.1}&\textbf{27.1}\\
    \bottomrule
    \end{tabular}
    \vspace{-10pt}
    \label{tab:ab_loss}
\end{table}

\begin{figure}[t!]
    \centering
    \includegraphics[width=0.98\linewidth]{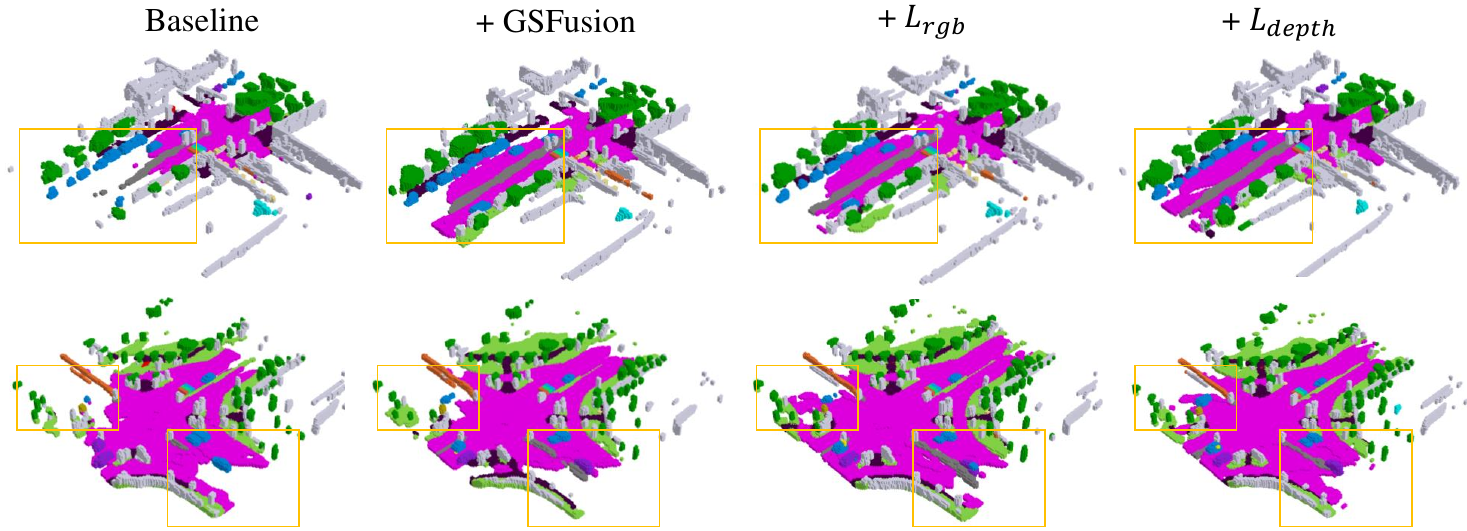}
    \caption{The visualization of ablation study on the impact of different components in nuScenes validation set.}
    \label{fig:viz_ab}
    \vspace{-16pt}
\end{figure}
 \vspace{-5pt}
\subsection{Optimization}
\label{loss}
Based above of our methods above, there are four main terms in our loss function.

\noindent\textbf{Occupancy Loss.}
we use cross-entropy loss $\mathcal{L}_{ce}$ and lovasz-softmax loss $\mathcal{L}_{ls}$ to supervise between predicted semantic occupancy $O$ and ground truth occupancy $\hat{O}$ as:
\begin{equation}
    \mathcal{L}_{occ} = \mathcal{L}_{ce} + \mathcal{L}_{ls}.
\end{equation}

\noindent\textbf{Explicit Depth Loss.}
In the 2D to 3D view transform, we will use a Depth Net~\cite{philion2020lift} to learn the depth with $\mathcal{L}_{d}$.

\noindent\textbf{Rendering Color Regularization.}
The color loss is calculated by MSE Loss between rendered color maps $\hat{I'}$ and the input camera multi-view images $I$ as:
{\setlength\abovedisplayskip{2pt}
\setlength\belowdisplayskip{2pt}
\begin{equation}
    \mathcal{L}_{rc} = \lambda_{rc}\cdot \left\|\hat{I'} - I\right\|_2,
\end{equation}
where $\lambda_{color}$ is a hyperparameter of $\mathcal{L}_{color}$.

\noindent\textbf{Rendering Depth Regularization.}
The ground truth depth maps are obtained by projecting paired LiDAR points onto images plane as $D$, and we use L1 loss to calculate $\mathcal{L}_{rd}$:
{\setlength\abovedisplayskip{2pt}
\setlength\belowdisplayskip{2pt}
\begin{equation}
    \mathcal{L}_{rd} = \lambda_{rd} \cdot \left\|\hat{D'} - D\right\|_1.
\end{equation}
where $\lambda_{rd}$ is a hyperparameter of $\mathcal{L}_{rd}$.

The overall objective function is: 
\begin{equation}
    \mathcal{L} = \mathcal{L}_{occ} + \mathcal{L}_{d} + \mathcal{L}_{rc} + \mathcal{L}_{rd}.
\end{equation}

\noindent\textbf{Discussion on different modal branch.} In Fig.~\ref{fw}, our LiDAR-camera fusion branch utilizes GSFusion and associated losses. In the LiDAR branch, depth regularization enhances LiDAR feature consistency without using GSFusion. In the camera branch, both color and depth regularization are applied with ground truth depth input. Without ground truth depth, only color regularization is employed.

\section{EXPERIMENT RESULTS}
\subsection{Implementation Details and Dataset}
\noindent\textbf{Implementation Details}.
We evaluate 3D semantic occupancy performance on the nuScenes occupancy validation set~\cite{caesar2020nuscenes, wei2023surroundocc} and the SemanticKITTI semantic scene completion set~\cite{behley2019semantickitti}. For the camera branch, we use ResNet50 or ResNet101~\cite{he2016deep} with FPN~\cite{lin2017feature} for both datasets. The view transformer~\cite{philion2020lift} generates a 3D feature volume of size $100\times100\times8$ and $128\times128\times16$, with 128 channels for nuScenes and SemanticKITTI, respectively. In the LiDAR branch, we voxelize 10 LiDAR sweeps and employ a voxel encoder for the nuScenes datasets. LiDAR features are augmented by selecting the 2 nearest neighborhoods from camera features. We use the same occupancy decoder and head as in~\cite{wang2023openoccupancy} with a cascade ratio of 2 for refining predictions. Our implicit volume rendering regularization incorporates two MLP networks as the density head and color head. The color head consists of three MLP layers, while the density head consists of either one MLP layer or a linear layer. The model is trained for 24 epochs on the nuScenes dataset and 30 epochs on the SemanticKITTI dataset, using a batch size of 8 across 8 A40 GPUs. We employ the AdamW~\cite{loshchilov2017decoupled} optimizer with a weight decay of 0.01 and an initial learning rate of 1e-4, along with a multi-step learning rate scheduler.

\noindent\textbf{NuScenes Dataset.}
nuScenes\cite{caesar2020nuscenes} is a large-scale autonomous driving benchmark including 1K driving scenarios, including 700 scenes for training, 150 scenes for validations, and 150 scenes for testing.  The 3D occupancy ground
truth provided by~\cite{wei2023surroundocc} of each sample with a voxel size of
$[0.5m,0.5m,0.5m]$ and represents $[200, 200, 16]$ dense voxel grids, which have 17 classes (16 semantics, 1 free).

\noindent\textbf{SemanticKITTI Dataset.}
The SemanticKITTI dataset~\cite{behley2019semantickitti} focuses on the semantic scene understanding with LiDAR point clouds and camera images. 
Specifically, the ground truth semantic occupancy is represented as the $[256, 256, 32]$ voxel grids. 
Each voxel is $[0.2m, 0.2m, 0.2m]$ large and annotated with 21 semantic classes (19 semantics, 1 free, 1 unknown). 
Following~\cite{cao2022monoscene, cheng2021s3cnet}, the 22 sequences are split into 10 sequences, 1 sequence, and 11 sequences for training, validation, and testing.

\subsection{Main Results}
To ensure fair comparisons, all results are either implemented by their authors or reproduced using official codes.

\begin{table*}[t!]
		\footnotesize
		\setlength{\tabcolsep}{0.004\linewidth}
		\caption{\textbf{3D Semantic occupancy prediction results on SemanticKITTI test set.} The C and L denote camera and LiDAR.}
		\vspace{-3mm}
		\newcommand{\classfreq}[1]{{~\tiny(\semkitfreq{#1}\%)}}  %
		\centering
		\begin{tabular}{l|c|c| c c c c c c c c c c c c c c c c c c c}
			\toprule
			Method
                & \makecell[c]{Modality}
			 & mIoU
			& \rotatebox{90}{\textcolor{road}{$\blacksquare$} road}
			\rotatebox{90}{\ \ \ \classfreq{road}} 
			& \rotatebox{90}{\textcolor{sidewalk}{$\blacksquare$} sidewalk}
			\rotatebox{90}{\ \ \ \classfreq{sidewalk}}
			& \rotatebox{90}{\textcolor{parking}{$\blacksquare$} parking}
			\rotatebox{90}{\ \ \ \classfreq{parking}} 
			& \rotatebox{90}{\textcolor{other-ground}{$\blacksquare$} other-grnd}
			\rotatebox{90}{\ \ \ \classfreq{otherground}} 
			& \rotatebox{90}{\textcolor{building}{$\blacksquare$}  building}
			\rotatebox{90}{\ \ \ \classfreq{building}} 
			& \rotatebox{90}{\textcolor{car}{$\blacksquare$}  car}
			\rotatebox{90}{\ \ \ \classfreq{car}} 
			& \rotatebox{90}{\textcolor{truck}{$\blacksquare$}  truck}
			\rotatebox{90}{\ \ \ \classfreq{truck}} 
			& \rotatebox{90}{\textcolor{bicycle}{$\blacksquare$}  bicycle}
			\rotatebox{90}{\ \ \ \classfreq{bicycle}} 
			& \rotatebox{90}{\textcolor{motorcycle}{$\blacksquare$} motorcycle}
			\rotatebox{90}{\ \ \ \classfreq{motorcycle}} 
			& \rotatebox{90}{\textcolor{other-vehicle}{$\blacksquare$}  other-veh.}
			\rotatebox{90}{\ \ \  \classfreq{othervehicle}} 
			& \rotatebox{90}{\textcolor{vegetation}{$\blacksquare$} vegetation}
			\rotatebox{90}{\ \ \ \classfreq{vegetation}} 
			& \rotatebox{90}{\textcolor{trunk}{$\blacksquare$}  trunk}
			\rotatebox{90}{\ \ \ \classfreq{trunk}} 
			& \rotatebox{90}{\textcolor{terrain}{$\blacksquare$} terrain}
			\rotatebox{90}{\ \ \ \classfreq{terrain}} 
			& \rotatebox{90}{\textcolor{person}{$\blacksquare$}  person}
			\rotatebox{90}{\ \ \ \classfreq{person}} 
			& \rotatebox{90}{\textcolor{bicyclist}{$\blacksquare$}  bicyclist}
			\rotatebox{90}{\ \ \ \classfreq{bicyclist}} 
			& \rotatebox{90}{\textcolor{motorcyclist}{$\blacksquare$}  motorcyclist.}
			\rotatebox{90}{\ \ \ \classfreq{motorcyclist}} 
			& \rotatebox{90}{\textcolor{fence}{$\blacksquare$} fence}
			\rotatebox{90}{\ \ \ \classfreq{fence}} 
			& \rotatebox{90}{\textcolor{pole}{$\blacksquare$} pole}
			\rotatebox{90}{\ \ \ \classfreq{pole}} 
			& \rotatebox{90}{\textcolor{traffic-sign}{$\blacksquare$} traf.-sign}
			\rotatebox{90}{\ \ \ \classfreq{trafficsign}} 
			\\
			\midrule

    	MonoScene~\cite{cao2022monoscene}&C  & 11.1 & 54.7 & 27.1 &         24.8 & 5.7 & 14.4 & 18.8 & 3.3 & 0.5 & 0.7 & 4.4& 14.9 & 2.4 & 19.5       & 1.0 & 1.4 & 0.4 & 11.1 & 3.3 & 2.1  \\
			
		SurroundOcc~\cite{wei2023surroundocc}&C  & 11.9 & 56.9 & 28.3 & 
            30.2 & 6.8 & 15.2 & 20.6 & 1.4 & 1.6 & 1.2 & 4.4 & 14.9 & 3.4 & 19.3 & 1.4 & 2.0 & 0.1 & 11.3& 3.9 & 2.4 \\ 
            
            OccFormer~\cite{zhang2023occformer} & C  &12.3& 55.9& 30.3& 31.5& 6.5 &15.7 &21.6& 1.2& 1.5& 1.7 &3.2& 16.8& 3.9 &21.3& 2.2& 1.1& 0.2 &11.9 &3.8 &3.7\\
            RenderOcc~\cite{pan2023renderocc} & C & 12.8& 57.2 & 28.4 & 16.1& 0.9& 18.2& 24.9&6.0& 3.1& 0.28 & 3.6& 26.2& 4.8& 3.6& 1.9& 3.3 & 0.3& 9.1 & 6.2& 3.3\\
            LMSCNet~\cite{roldao2020lmscnet} &L &17.0&64.0& 33.1 &24.9& 3.2& 38.7& 29.5 &2.5& 0.0& 0.0& 0.1& 40.5& 19.0 &30.8& 0.0& 0.0& 0.0& 20.5& 15.7& 0.5\\
            JS3C-Net~\cite{yan2021sparse}&L  &23.8& 64.0& 39.0 & 34.2 & \textbf{14.7}& 39.4  &33.2 & \textbf{7.2} & \textbf{14.0} & \textbf{8.1}& \textbf{12.2}  &43.5&19.3 & 39.8 &\textbf{ 7.9}  &\textbf{5.2} & 0.0 & 30.1 & 17.9 &15.1   \\
            SSC-RS~\cite{mei2023ssc}&L& 24.2&73.1& \textbf{44.4}& 38.6 &17.4 &\textbf{44.6} &36.4 &5.3 &10.1 &5.1& 11.2& \textbf{44.1} &26.0 &41.9 &4.7& 2.4& 0.9& 30.8& 15.0& 7.2\\
            M-CONet~\cite{wang2023openoccupancy}&C\&L  & 20.4 & 60.6 & 36.1 &29.0  & 13.0 & 38.4 &33.8 & 4.7 &3.0 &2.2  & 5.9 & 41.5 &20.5 &35.1 & 0.8 & 2.3 & \textbf{0.6} & 26.0 &18.7 & 15.7  \\
            \midrule
            
            
            Co-Occ (Ours)&C\&L &\textbf{24.4}  &\textbf{72.0}  &43.5 &\textbf{42.5}  & 10.2 &35.1  & \textbf{40.0}& 6.4&4.4&3.3&8.8&41.2&\textbf{30.8}&\textbf{40.8} & 1.6 & 3.3& 0.4 & \textbf{32.7}& \textbf{26.6}& \textbf{20.7}   \\
			\bottomrule
		\end{tabular}
		
		\vspace{1mm}
		
		\label{tab:kitti}
		
	\end{table*}

\begin{figure*}[t!]
    \centering
    \includegraphics[width=0.85\linewidth]{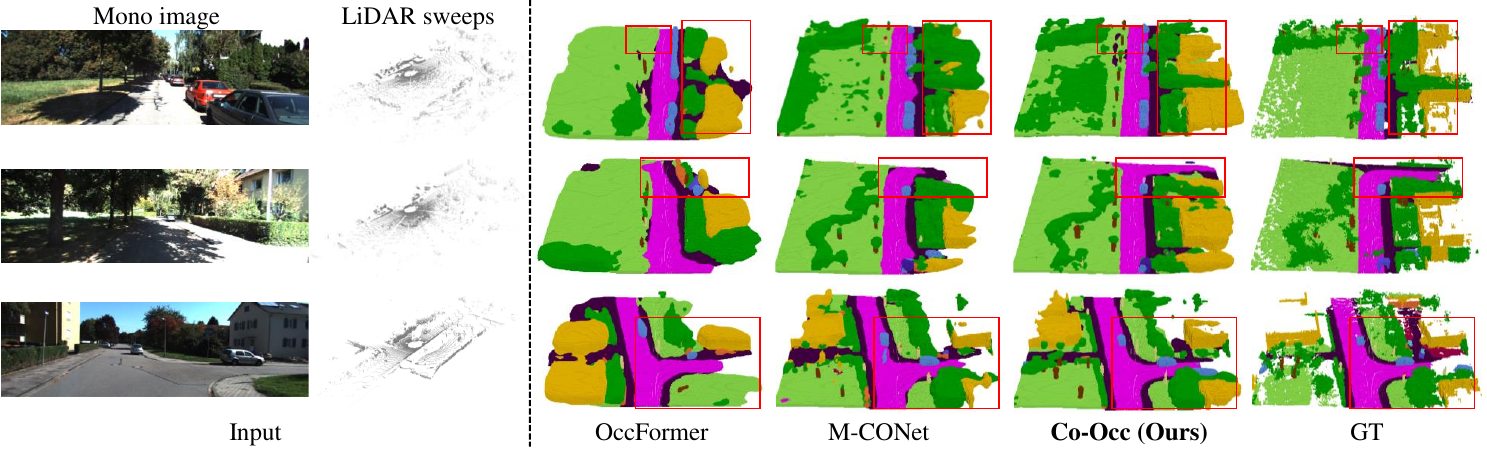}
    \vspace{-10pt}
    \caption{The qualitative comparisons results on SemanticKITTI validation set. The input monocular image and LiDAR sweeps are shown on the left and the 3D semantic occupancy results from OccFormer~\cite{zhang2023occformer} (OccFormer predicts results using only mono image), M-CONet~\cite{wang2023openoccupancy}, our Co-Occ, and the annotations are then visualized sequentially. \textbf{Better viewed when zoomed in.}}
    \label{fig:com_kitti}
    \vspace{-18pt}
\end{figure*}

\noindent\textbf{Results on NuScenes Dataset:}
We first conduct experiments on the nuScenes~\cite{caesar2020nuscenes} dataset and compare with several SOTA methods with different modalities, \ie, camera-only methods~\cite{cao2022monoscene, li2022bevformer, zhang2023occformer, wei2023surroundocc, wang2023openoccupancy, li2023fb, pan2023renderocc}, LiDAR-only methods~\cite{roldao2020lmscnet, wang2023openoccupancy}, and LiDAR-camera fusion methods~\cite{wang2023openoccupancy}. 
The settings for MonoScene~\cite{cao2022monoscene}, BEVFormer~\cite{li2022bevformer}, and SurroundOcc~\cite{wei2023surroundocc} involve an input size of $900\times 1600$ and utilize a ResNet101-DCN backbone, consistent with the test settings used in the previous work~\cite{wei2023surroundocc}. As for OccFormer~\cite{zhang2023occformer}, C-CONet~\cite{wang2023openoccupancy}, M-CONet~\cite{wang2023openoccupancy}, FB-Occ~\cite{li2023fb}, and RenderOcc\cite{pan2023renderocc}, we combine the multi-camera or LiDAR-camera occupancy predictions and voxelize them with a voxel size of $0.5m$, matching our setting where the input image size is $896\times1600$ and the 2D backbone is ResNet101.
Tab.~\ref{tab:nusc} demonstrates that our Co-Occ with LiDAR and the camera achieves a notable increase of \textbf{2.4\%} mIoU compared to M-CONet~\cite{wang2023openoccupancy}, which also utilizes both camera and LiDAR data, thus showcasing the effectiveness of our fusion techniques. Besides, our methods utilizing the camera-only branch show improvements in both mIoU and IoU with the same input size and 2D backbone methods~\cite{zhang2023occformer, wang2023openoccupancy, li2023fb, pan2023renderocc}. Additionally, our Co-Occ with the LiDAR-only branch also demonstrates significant improvement~\cite{roldao2020lmscnet, wang2023openoccupancy}. It is important to note that the IoU score of LiDAR-only is better than the LiDAR-camera method, suggesting that the LiDAR sweeps more regions but with less accurate categorization for the semantic classes in the context of LiDAR-camera fusion methods.
Furthermore, as shown in Fig.~\ref{fig:com_nusc}, the qualitative results illustrate that our methods predict a more fine-grained and accurate 3D semantic occupancy.

\noindent\textbf{Results on SemanticKITTI Dataset:}
To further validate the effectiveness of our techniques, we conducted a comparative analysis with single-modal and multi-modal state-of-the-art methods~\cite{cao2022monoscene, wei2023surroundocc, zhang2023occformer, roldao2020lmscnet, yan2021sparse, wang2023openoccupancy} on the SemanticKITTI test set~\cite{behley2019semantickitti}. 
As shown in Tab.~\ref{tab:kitti}, our methods outperform JS3CNet~\cite{yan2021sparse} by 0.6\% mIoU and SSC-RS~\cite{mei2023ssc} by 0.2\% mIoU, even though they utilize additional LiDAR segmentation supervision.
Furthermore, our methods show a significant improvement of \textbf{4.0\%} mIoU over M-CONet~\cite{wang2023openoccupancy}. Fig.~\ref{fig:com_kitti} depicts the qualitative results on SemanticKITTI validation, indicating our superior performance in both scene occupancy complementary and accuracy of object details. These results further confirm the effectiveness of our techniques.


\subsection{Ablation Studies}

\begin{table}[t!]
    \footnotesize
    \centering
    \caption{Ablation study on fusion strategy and parameter selection. }
    \setlength{\tabcolsep}{0.05\linewidth}
    \begin{tabular}{c|cc|cc}
    \toprule
    Fusion Method&k & $n_s$ & IoU$\uparrow$ & mIoU$\uparrow$ \\ \midrule
    \multirow{4}{*}{GSFusion}&1&56& 39.9& 25.9\\
    &1&112& 40.5& 26.3\\ 
    &2&112& \underline{41.1} & \underline{27.1}\\
    &3&112& \textbf{41.3} & \textbf{27.2}\\  
    Concatenated &2&112& 38.2 &24.5\\
    Weighted~\cite{wang2023openoccupancy} &2&112& 39.3& 25.6\\
    \bottomrule
    \end{tabular}
    \vspace{-10pt}
    \label{tab:ab:param}
\end{table}

\noindent \textbf{Ablation on Architectural Components}.
As depicted in Tab.~\ref{tab:ab_loss} and Fig.~\ref{fig:viz_ab}, our GSFusion module exhibits a 1.7\% mIoU improvement compared to our baseline, which employs the concatenation fusion between LiDAR and camera features. Furthermore, incorporating the implicit volume rendering regularization through the inclusion of $\mathcal{L}_{rc}$ and $\mathcal{L}_{rd}$ losses leads to additional performance enhancement in our method. This showcases the effectiveness of each module and the loss in our approach.
     
 

\noindent \textbf{Ablation on Different Fusion Strategies.}
In Tab.~\ref{tab:ab:param}, we conduct an ablation study on fusion strategies, comparing concatenated fusion and weighted fusion, \eg,~\cite{wang2023openoccupancy}, used in previous LiDAR-camera fusion-based works. We find that straight concatenated fusion methods underperform compared to other fusion techniques. This may be due to limited utilization of the inaccurate calibration of the two modalities. In contrast, our GSFusion strategy leverages both geometric and semantic information, leading to a significant 1.5\% improvement over weighted fusion. We analyze that although weighted fusion methods use a network to adaptively learn the fusion weights of camera and LiDAR features, inaccurate calibrations among LiDAR and camera features may still lead to errors in the weighted fusion process.

\noindent \textbf{Ablation on Parameter Selection.}
We perform an internal ablation study on our two components, focusing on the selection of the numbers for the KNN strategies and the sampling points.

\begin{table}[t!]
    \tabcolsep=0.15cm
    \footnotesize
    \centering
    \caption{Efficiency analysis on Co-Occ on a single RTX A40 GPU.}
    \vspace{-5pt}
    \begin{tabular}{cccccc}
    \toprule
    Image Size& 2D backbone &IoU & mIoU& Memory (G)&Latency (s)\\ \midrule
    $256\times 704$  & R50 & 40.9& 25.0&10.75&0.45\\
    $896\times 1600$ & R50 & \textbf{41.4}& 26.7&11.65&0.55\\
    $896\times 1600$& R101 & 41.1& \textbf{27.1}&11.78&0.58\\
    \bottomrule
    \end{tabular}
    \vspace{-10pt}
    \label{tab:ab:efficiency}
    
\end{table}
\begin{table}[t!]
    \tabcolsep=0.15cm
    \footnotesize
    \centering
    \caption{3D semantic occupancy results with different ranges.}
    \vspace{-5pt}
    \begin{tabular}{lcccccc}
    \toprule
    \multirow{2}{*}{Method}&\multicolumn{3}{c}{IoU}& \multicolumn{3}{c}{mIoU}\\ \cline{2-7}
     & 25m & 50m& 100m & 25m& 50m & 100m\\ \midrule
    M-CONet~\cite{wang2023openoccupancy}&60.9 & 51.0&39.2 &36.9 &31.8 &24.7\\
    Co-Occ (ours)& 62.7& 53.0&41.1 & 40.3&34.6 &27.1\\
    Improvements (\%)& +1.8& +2.0&+2.0 & +3.4&+2.8 &+2.4\\
    \bottomrule
    \end{tabular}
    \vspace{-15pt}
    \label{tab:ab:ranges}
    
\end{table}

\noindent  1) The numbers of neighbors $k$:
As shown in Tab.~\ref{tab:ab:param}, we notice a correlation between performance and the number of selected neighbors. Selecting two nearest neighbors improves performance by approximately 0.8\% mIoU to select one nearest neighbor. However, selecting three neighbors does not provide an obvious enhancement and adds a computational burden. Thus, we choose $k=2$ as our optimal number.

\noindent 2) The numbers of sampling points $n_s$:
We conduct this study using the same depth bound and half depth bound as the sampling point parameters as shown in Tab.~\ref{tab:ab:param}. After careful evaluation, we determined that setting the sampling parameter to 112 yields the best performance.

\noindent\textbf{Analysis of model efficiency.}
Tab.~\ref{tab:ab:efficiency} evaluates the inference time and memory usage for different image sizes and 2D backbones in our LiDAR-camera fusion-based branch. These experiments were conducted on a single RTX A40 GPU. A model with low image resolution and ResNet50 exhibits reduced computational costs and latency, but its performance is limited. Increasing the image size and depth of ResNet does not significantly increase memory usage or latency, but it leads to a significant improvement in performance.

\noindent\textbf{Analysis of results within different ranges.}
We further propose to evaluate different ranges surrounding the car comprehensively. 
We present the IoU and mIoU data separately for the volumes of $25m 
\times 25m \times 8m$, $50m \times 50m \times 8m$, and $100m \times 100m \times 8m$. Understanding short-range areas is crucial as it allows less time for autonomous vehicles to react. As depicted in Tab.~\ref{tab:ab:ranges}, our method shows substantial improvements (3.4\%) over M-CONet~\cite{wang2023openoccupancy} in short-range areas. 
Due to the sparsity of LiDAR sweeps in long-range areas, only a few pixels determine the depth of a large area. Despite the diminishing improvements of our methods over M-CONet~\cite{wang2023openoccupancy}, it still maintains a 2.4\% improvement in mIoU scores, demonstrating the effectiveness of our methods across different ranges.

\vspace{-5pt}
\section{CONCLUSION}
In this letter, we proposed a multi-modal 3D semantic occupancy prediction framework that combines explicit GSFusion and implicit volume rendering regularization. Our approach improves the interaction between LiDAR and camera data and enables dense semantic occupancy predictions. We introduced a KNN-based GSFusion method for optimal feature fusion in the voxel space. Besides, we incorporated implicit volume rendering regularization to enhance the fused representation by connecting 2D images and 3D LiDAR sweeps. Extensive experiments confirmed the effectiveness of our proposed method.





\ifCLASSOPTIONcaptionsoff
  \newpage
\fi

\section{Appendix}
\subsection{Details}
\subsubsection{The Voxel Interaction of LiDAR and Camera}
Fig.~\ref{fig:suppl_feat} (a) illustrates the challenge of aligning images and point clouds caused by inaccurate extrinsic parameters. Direct geometric alignment is difficult to achieve. To address the accumulation of errors resulting from misalignment, we propose GSFusion. This method searches for nearby features to ensure both geometric and semantic alignment, enabling each LiDAR voxel feature to interact with K neighboring lifted pixel features in the fusion process. This expands the perception field, allowing for a more comprehensive and robust fusion of image and point features.
Furthermore, Fig.~\ref{fig:suppl_feat} (b) highlights the impact of the sparsity of LiDAR point clouds on voxel interaction with the camera. To address this, the rendering process ensures dense representations for LiDAR features, camera features, or LiDAR-camera features, as depicted in Fig.~\ref{fig:suppl_feat} (c). This ensures sufficient voxel interaction and improves overall performance.
\begin{figure}[h]
    \centering
    \includegraphics[width=\linewidth]{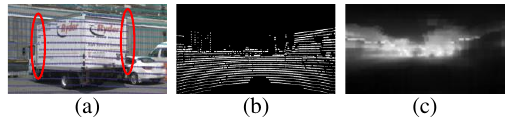}
    \caption{(a) illustrates the inaccurate calibrations between the projected points and the corresponding images.
(b) displays the sparse distribution of projected LiDAR points.
(c) showcases the rendered depth obtained from the fusion of LiDAR-camera features.}
    \label{fig:suppl_feat}
\end{figure}

\begin{table*}[t]
\setlength{\tabcolsep}{0.0045\linewidth}
\newcommand{\classfreq}[1]{{~\tiny(\semkitfreq{#1}\%)}}  %
\centering
\caption{\textbf{3D Semantic occupancy prediction results on nuScenes-Occupancy validation set.} We report the geometric metric IoU, semantic metric mIoU, and the IoU for each semantic class. The C, D, and L denotes camera, depth, and LiDAR, respectively. \textbf{Bold} represents the best score.}
\resizebox{.85\linewidth}{!}{
\begin{tabular}{l|c | c c | c c c c c c c c c c c c c c c c}
    \toprule
    Method
    & \makecell[c]{Modality}
    & IoU& mIoU
    & \rotatebox{90}{\textcolor{nbarrier}{$\blacksquare$} barrier}
    & \rotatebox{90}{\textcolor{nbicycle}{$\blacksquare$} bicycle}
    & \rotatebox{90}{\textcolor{nbus}{$\blacksquare$} bus}
    & \rotatebox{90}{\textcolor{ncar}{$\blacksquare$} car}
    & \rotatebox{90}{\textcolor{nconstruct}{$\blacksquare$} const. veh.}
    & \rotatebox{90}{\textcolor{nmotor}{$\blacksquare$} motorcycle}
    & \rotatebox{90}{\textcolor{npedestrian}{$\blacksquare$} pedestrian}
    & \rotatebox{90}{\textcolor{ntraffic}{$\blacksquare$} traffic cone}
    & \rotatebox{90}{\textcolor{ntrailer}{$\blacksquare$} trailer}
    & \rotatebox{90}{\textcolor{ntruck}{$\blacksquare$} truck}
    & \rotatebox{90}{\textcolor{ndriveable}{$\blacksquare$} drive. suf.}
    & \rotatebox{90}{\textcolor{nother}{$\blacksquare$} other flat}
    & \rotatebox{90}{\textcolor{nsidewalk}{$\blacksquare$} sidewalk}
    & \rotatebox{90}{\textcolor{nterrain}{$\blacksquare$} terrain}
    & \rotatebox{90}{\textcolor{nmanmade}{$\blacksquare$} manmade}
    & \rotatebox{90}{\textcolor{nvegetation}{$\blacksquare$} vegetation} \\
    \midrule
    MonoScene~\cite{cao2022monoscene} & C   & 18.4 & 6.9 & 7.1  & 3.9  &  9.3 &  7.2 & 5.6  & 3.0  &  5.9& 4.4& 4.9 & 4.2 & 14.9 & 6.3  & 7.9 & 7.4  & 10.0 & 7.6 \\
  
    TPVFormer~\cite{huang2023tri} &C &  15.3 &  7.8 & 9.3  & 4.1  &  11.3 &  10.1 & 5.2  & 4.3  & 5.9 & 5.3&  6.8& 6.5 & 13.6 & 9.0  & 8.3 & 8.0  & 9.2 & 8.2 \\

        3DSketch~\cite{chen20203d} &  C\&D & 25.6 & 10.7  & 12.0 &  5.1 &  10.7 &  12.4 & 6.5  & 4.0  & 5.0 & 6.3&  8.0&  7.2& 21.8 &  14.8 & 13.0 &  11.8 & 12.0 & 21.2 \\
        
        AICNet~\cite{li2020anisotropic} & C\&D   &   23.8 & 10.6  & 11.5  & 4.0  & 11.8  & 12.3&  5.1 & 3.8  & 6.2  & 6.0 & 8.2&  7.5&  24.1 & 13.0 & 12.8  & 11.5 & 11.6  &  20.2\\

        LMSCNet~\cite{roldao2020lmscnet} & L &   27.3 & 11.5 & 12.4&  4.2 & 12.8  & 12.1  & 6.2  &  4.7 & 6.2 & 6.3&  8.8&  7.2& 24.2 & 12.3  & 16.6 & 14.1  & 13.9 & 22.2 \\

    JS3C-Net~\cite{yan2021sparse} &L &   30.2  & 12.5 & 14.2 & 3.4  & 13.6  & 12.0  & 7.2  &  4.3 & 7.3 & 6.8&  9.2& 9.1 & 27.9 & 15.3  & 14.9 & 16.2  & 14.0 & 24.9 \\
    
        M-CONet~\cite{wang2023openoccupancy} & C\&L &   29.5 & 20.1 &  23.3  & 13.3& 21.2 & 24.3& \textbf{15.3}  & 15.9& 18.0 & 13.3 & 15.3 & 20.7 & 33.2 & 21.0 & 22.5  & 21.5 &19.6 & 23.2  \\ \midrule
        Co-Occ (Ours) & C\&L &\textbf{30.6}&\textbf{21.9}&\textbf{26.5}  &\textbf{16.8}&\textbf{22.3}&\textbf{27.0}&10.1&\textbf{20.9}&\textbf{20.7} &\textbf{14.5 }&\textbf{16.4} &\textbf{21.6}&\textbf{36.9}&\textbf{23.5}&\textbf{25.5} &\textbf{23.7} &\textbf{20.5} &\textbf{23.5} \\
    \bottomrule

\end{tabular}}

\label{tab:suppl_nusc}
\end{table*}

\begin{figure*}[h]
    \centering
    \includegraphics[width=\linewidth]{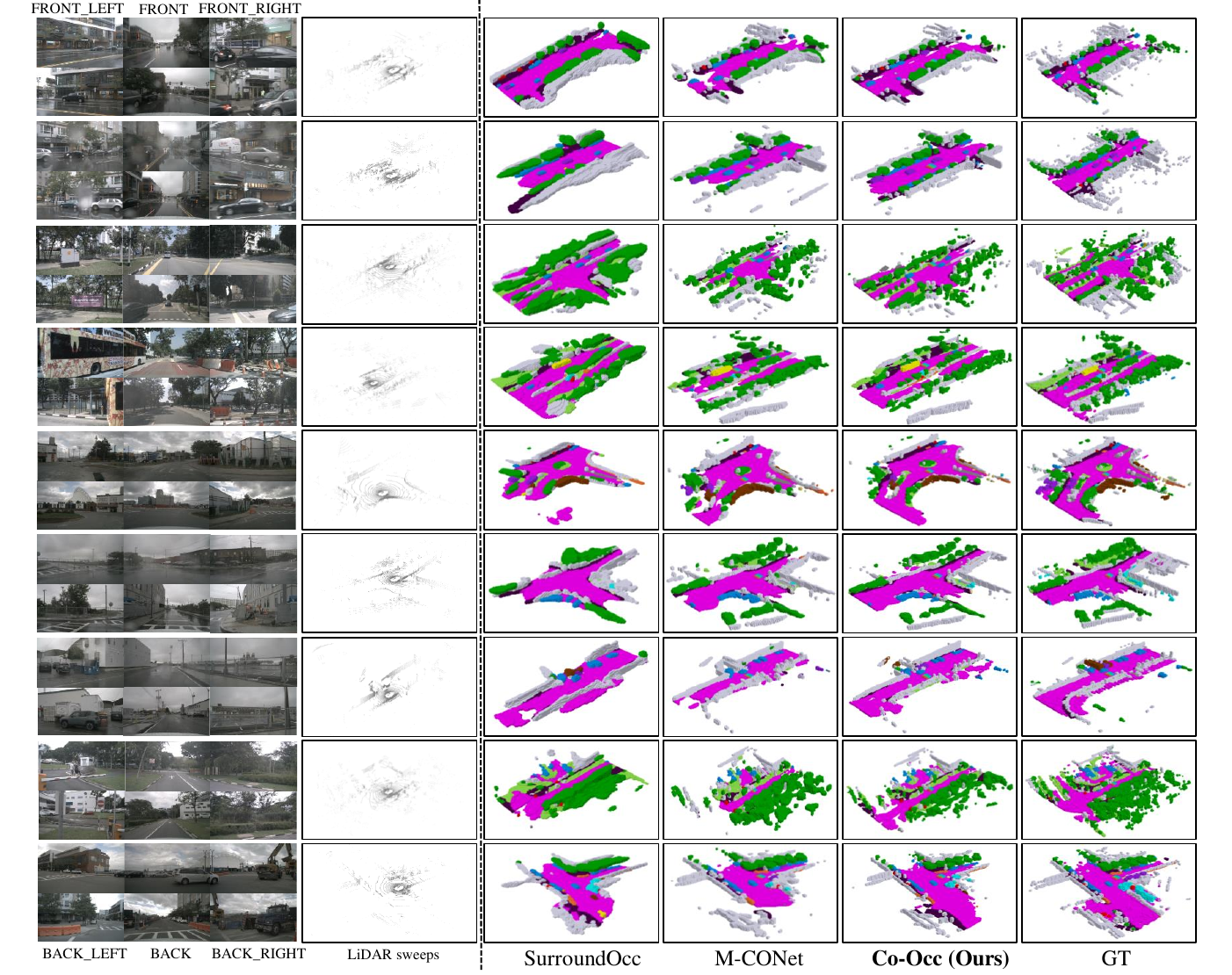}
    \caption{The additional qualitative comparisons results on nuScenes validation set~\cite{behley2019semantickitti}. The leftmost column shows the input surrounding images and LiDAR sweeps, the following three columns visualize the 3D semantic occupancy prediction from SurroundOcc~\cite{wei2023surroundocc} (SurroundOcc predicts results using only cameras), M-CONet~\cite{wang2023openoccupancy}, our Co-Occ, and the annotation from~\cite{wei2023surroundocc}. \textbf{Better viewed when zoomed in.}}
    \label{fig:suppl_nusc}
\end{figure*}

\begin{figure*}[h]
    \centering
    \includegraphics[width=\linewidth]{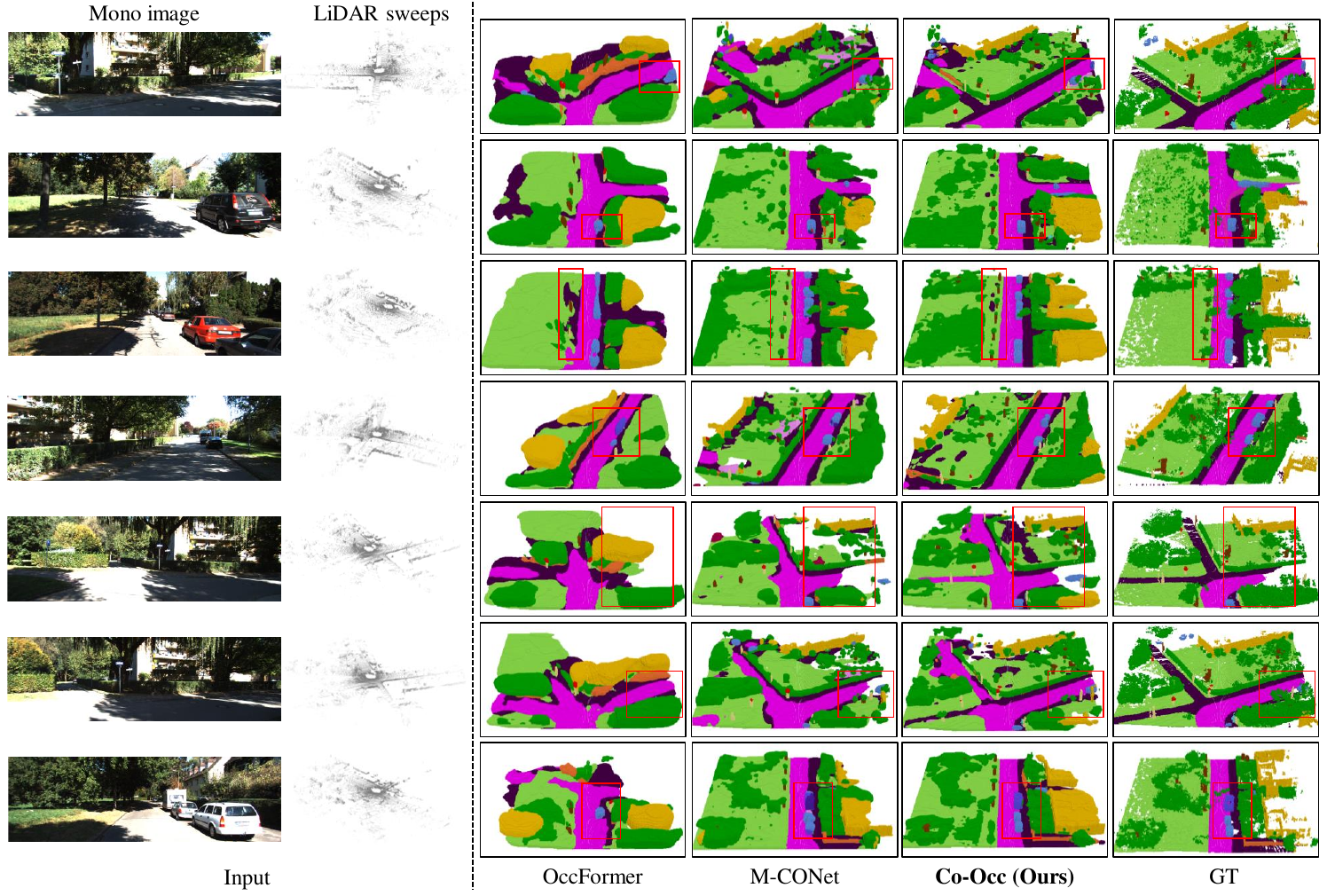}
    \caption{The additional qualitative comparisons results on SemanticKITTI validation set. The input monocular image and LiDAR sweeps are shown on the left and the 3D semantic occupancy results from OccFormer~\cite{zhang2023occformer} (OccFormer predicts results using only mono image), M-CONet~\cite{wang2023openoccupancy}, our Co-Occ, and the annotations are then visualized sequentially. \textbf{Better viewed when zoomed in.}}
    \label{fig:suppl_kitti}
\end{figure*}
\subsubsection{Other Implementation Details}
In our implicit volume rendering regularization, we employ two Multi-Layer Perceptron (MLP) networks as our density head and color head. These MLP networks are utilized to generate the density grid and color grid from the sampled frustum features. The color head consists of three layers of MLP, while the density head consists of either one layer of MLP or a Linear layer. This configuration strikes a balance between training memory requirements and rendering performance.
\subsection{Evaluation Metrics}
\noindent\textbf{IoU metrics.}
The Intersection over Union (IoU) is a metric used to determine whether a voxel is occupied or empty~\cite{caesar2020nuscenes, wang2023openoccupancy}. It treats all occupied voxels as the occupied class and all others as the empty class. The IoU is calculated as follows: 
\begin{equation} 
\rm{IoU} = \frac{TP_o}{TP_o+FP_o+FN_o}, 
\end{equation} 
where $TP_o, FP_o, FN_o$ represent the number of true positives, false positives, and false negatives of the occupied class, respectively.

\noindent\textbf{mIoU metrics.}
The mean Intersection over Union (mIoU) is a metric that calculates the average IoU for each semantic class. It is defined as follows:
\begin{equation}
    \rm{mIoU} = \frac{1}{N_c}\sum_{c=1}^{N_c}\frac{TP_c}{TP_c+FP_c+FN_c},
\end{equation}
where $TP_c, FP_c, FN_c$ represent the number of true positives, false positives, and false negatives for class $c$, respectively, and $N_c$ is the total number of classes.

\subsection{More Results}
\noindent\textbf{More Quantitative Results on NuScenes dataset.}
We supply the 3D semantic occupancy prediction results on different voxel resolutions. As the openoccupancy benchmark~\cite{wang2023openoccupancy}, the voxel resolution is $0.2m$ in a volume of $512\times512\times40$ for occupancy predictions. As presented in Tab.~\ref{tab:suppl_nusc}, we test other method on nuScenes-occupancy validation set~\cite{caesar2020nuscenes, wang2023openoccupancy}. our Co-Occ method obtain \textbf{1.8\%} mIoU improvement among the camera-LiDAR fusion-based M-CONet. Besides, the scores of most semantic classes have a large margin improvement. These experiments validate the effectiveness of our method that not only obatin better performance on one voxel resolution.

\noindent\textbf{More Visualization Results on NuScenes Dataset.}
Due to space constraints, we have included additional visual results in Fig. \ref{fig:suppl_nusc}. These results demonstrate that our method has more precise details compared to the camera-only method~\cite{wei2023surroundocc}, while also achieving greater consistency than other LiDAR-camera fusion-based methods~\cite{wang2023openoccupancy}. Furthermore, the effectiveness of our methods in challenging conditions is validated through the \textit{video demo}.

\noindent\textbf{More Visualization Results on SemanticKITTI Dataset.}
Similarly, we present additional qualitative results using the SemanticKITTI validation dataset in Fig.~\ref{fig:suppl_kitti}. Our method's semantic predictions clearly outperform not only in dynamic objects like cars but also in capturing the complementary aspects of road and vegetation. This showcases the effectiveness and versatility of our approach.

\noindent\textbf{Video Demo.}
To provide a comprehensive understanding of our method and showcase the dynamic performance of our results, we have prepared a video demo. This demo visually presents our continuous video visualization results, allowing you to observe our method in action and gain a deeper understanding of its process. 

{\small
\bibliographystyle{IEEEtran}
\bibliography{IEEEfull}
}








\end{document}